# SmilesT5: Domain-specific pretraining for molecular language models

Philip Spence[a,b], Brooks Paige[c,d] and Anne Osbourn[a]

[a]Department of Biochemistry and Metabolism, John Innes Centre, Norwich, UK
[b]HotHouse Therapeutics, Centrum, Norwich Research Park, Norwich, UK
[c]Centre for Artificial Intelligence, University College London, London, UK
[d]The Alan Turing Institute, UK

## Abstract

Molecular property prediction is an increasingly critical task within drug discovery and development. Typically, neural networks can learn molecular properties using graph-based, language-based or feature-based methods. Recent advances in natural language processing have highlighted the capabilities of neural networks to learn complex human language using masked language modelling. These approaches to training large transformer-based deep learning models have also been used to learn the language of molecules, as represented by simplified molecular-input line-entry system (SMILES) strings. Here, we present novel domain-specific text-to-text pretraining tasks that yield improved performance in six classification-based molecular property prediction benchmarks, relative to both traditional likelihood-based training and previously proposed fine-tuning tasks. Through ablation studies, we show that data and computational efficiency can be improved by using these domain-specific pretraining tasks. Finally, the pretrained embeddings from the model can be used as fixed inputs into a downstream machine learning classifier and yield comparable performance to finetuning but with much lower computational overhead.

## 1 Introduction

Molecular property prediction is a critical task in drug discovery and allows for rapid identification of bioactive molecules from large datasets [1, 2]. Recently, graph-based neural networks (GNNs) have gained traction and have shown great promise for learning molecular properties [3-5]. Yang et al. [6] published a directed message passing neural network (D-MPNN) model that substantially improved the performance of graph-based methods and was later used to aid in the discovery of novel antibiotics [7, 8]. In message passing neural networks (MPPNs) molecules are represented as graphs, with atoms as the nodes and bonds as the edges allowing for the features of neighbouring atoms to be passed between each other, capturing both the bond and atom features. While these models perform well on datasets containing a large chemical space for many applications of molecular property prediction there is a lack of labelled data, resulting in small datasets that only capture a small chemical space. To overcome this, models have been developed that are first pre-trained on a large dataset in a self-supervised fashion before being finetuned on a smaller dataset of interest [9]. In 2022, Fang et al. published the KANO model, which exploited a knowledge-graph of atom features and a contrastive pretraining task and showed significant improvement upon previous MPNN-based methods that lack pretraining [10].

An alternative to graph-based models are language-based models that typically learn molecules as SMILES strings, which contain all the information necessary to generate a molecular graph [11-13]. Typically, these models are pretrained on very large datasets using masked language modelling (MLM) and then finetuned on smaller datasets of interest. One such example is ChemBERTa, published in 2020 [14]. ChemBERTa is a Robustly Optimised Bidirectional Encoder Representation from Transformers (RoBERTa)-based model that was pretrained on the SMILES strings of molecules from the PubChem dataset using an MLM task [14]. The same authors then followed up with ChemBERTa-2, which was pretrained on a larger dataset and also included a separate model trained using multi-task regression (MTR) on 200 molecular features [15], both models performing competitively against common benchmarks. More recently, engineers at the International Business Machines Corporation (IBM) published MolFormer and MolFormer-XL, two language-based models trained on 100M and 1.1B molecules, respectively. Both models outperformed the ChemBERTa models in a number of different benchmarks [16], and are pretrained using an MLM task.

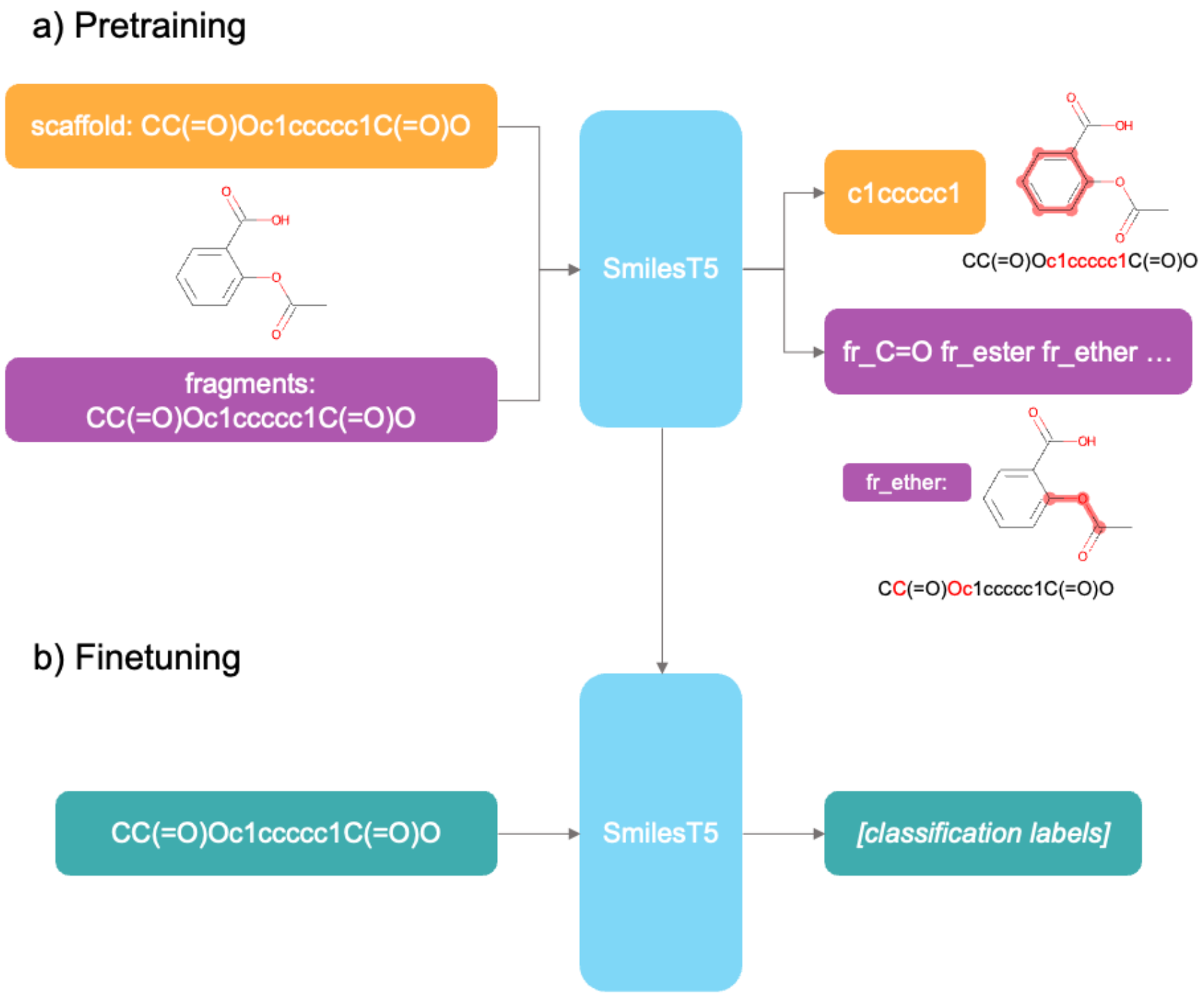

Figure 1: **Graphical depiction of the SmilesT5 model.** a) the pretraining tasks in which the model is asked to reconstruct the Murcko scaffold and pick out the presence of molecular fragments, and b) the pretrained model is then finetuned in a text-to-text tasks in which the model outputs the label token(s) as text

Masked language modelling has proven to be an effective pretraining task for many large language models within the natural language processing (NLP) domain [17, 18]. There have, however, been many different pretraining approaches developed within the context of NLP [19, 20]. In 2020, developers at Google published the Text-to-Text Transfer Transformer (T5) model which utilised a unique approach that reconfigured all natural language processing tasks into a text-to-text framework, allowing for multi-task learning [21]. This approach provided a simple framework for pretraining language models on many different tasks simultaneously and achieved state-of-the-art results in many benchmarks. Here, we present SmilesT5, a T5-based model pretrained on multiple text-to-text tasks that aim to capture knowledge specific to the domain of molecules (Figure 1). By combining this novel approach for learning the language of molecules and treating all classification tasks as text-to-text, SmilesT5 yields significantly improved performance on downstream molecular property prediction tasks.

# 2 Results & Discussion

## 2.1 Pretraining Tasks

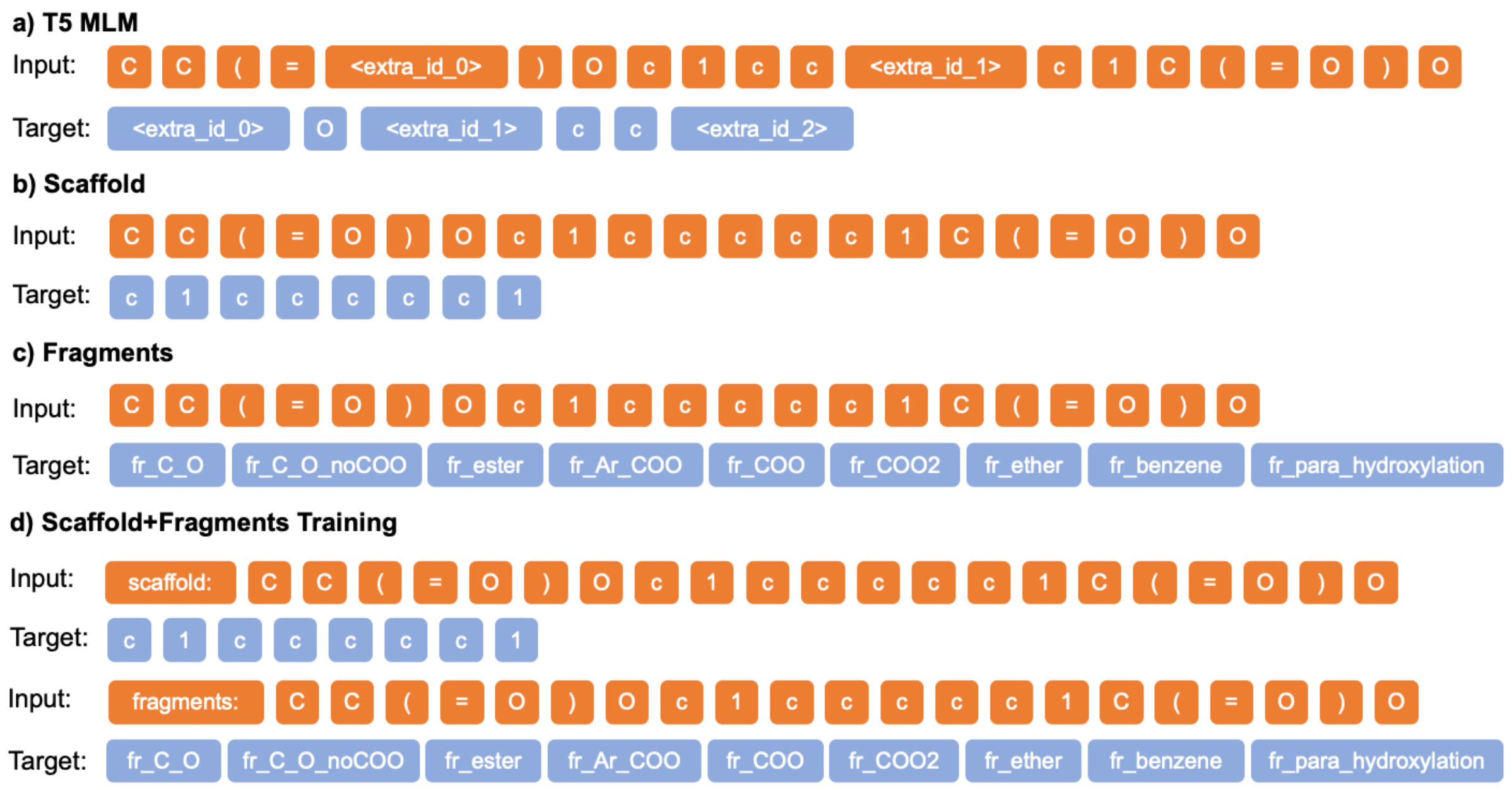

Figure 2: **Example of the tasks used to pretrain the SmilesT5 model.** a) The T5 masked language modelling task involves masking randomly sampled tokens from the input. b) Generating the Murcko scaffold of a molecule from the SMILES string (Scaffold task). c) Generating a text sequence of containing the names of the molecular fragments that are present in the molecule (Fragments task). d) By prepending the input task with either the "scaffold:" or "fragments:" token, the model can simultaneously learn the Scaffold and Fragments tasks, respectively (Scaffold+Fragments task).

To investigate the effect of different domain-specific pretraining tasks we trained four separate *small* T5 models (see Table 1, Methods, for model size definitions). The first model (MLM) was pretrained using the T5 MLM task (Figure 2a). Due to the text-to-text approach of the T5 model, MLM tasks are treated slightly differently to autoencoding models such as BERT, where a sample of tokens, typically 15%, are replaced with a mask token and the pretraining task is to rebuild the original sequence. For a T5 model, the original paper [21] outlines a new approach for denoising the original sequence where 15% of the tokens in the input are replaced by sentinel tokens, and the corresponding unmasked tokens in the target are also replaced. If the random sampling results in consecutive tokens being masked, then these will be replaced by a single sentinel token. For our MLM pretraining task, we sampled 15% of the inputs to be replaced by sentinel tokens.

The second model (Scaffold) was pretrained using the Murcko scaffold of the molecule as the target text (Figure 2b). The molecular scaffolds of molecules have been well studied and are one of the fundamental principles in structure-activity relationships [22, 23]. By learning to identify the tokens that compose the Murcko scaffold the model is required to learn the molecular structure of the molecule.

The third model (Fragments) was pretrained to predict the presence of defined molecular fragments in the molecule (Figure 2c). Molecular fragments and functional groups play a critical role in medicinal chemistry and have a substantial impact upon molecular properties [24, 25]. As with the Scaffold task, the fragments are not always contiguous across the SMILES string, especially in the case of fragments containing rings, and will therefore require the model to learn the underlying molecular structures.

The final model (Scaffold+Fragments) was pretrained using both the scaffold and fragment tasks simultaneously, where each input SMILES is duplicated and prepended by the task token "scaffold:" or "fragments:", respectively (Figure 2d). By performing both tasks simultaneously, the model will learn to capture important molecular structures and properties.

All four models were pretrained on 1M molecules from the zinc15 dataset [26], with 100K molecules being held back for validation using the scaffold splitting method outlined in MoleculeNet [2].

Against the test dataset, which is composed of molecules with unseen Murcko scaffolds, the Scaffold+Fragments pretraining task achieves a BLEU score of 0.989 and a word rate error of 0.125, with 98.7% accuracy of the predicted texts to the target texts. This demonstrates the ability of the model to accurately learn the underlying chemistry of SMILES strings, especially impressive is when the scaffold or fragments are not contiguous in the SMILES string (see SI Figure 1).

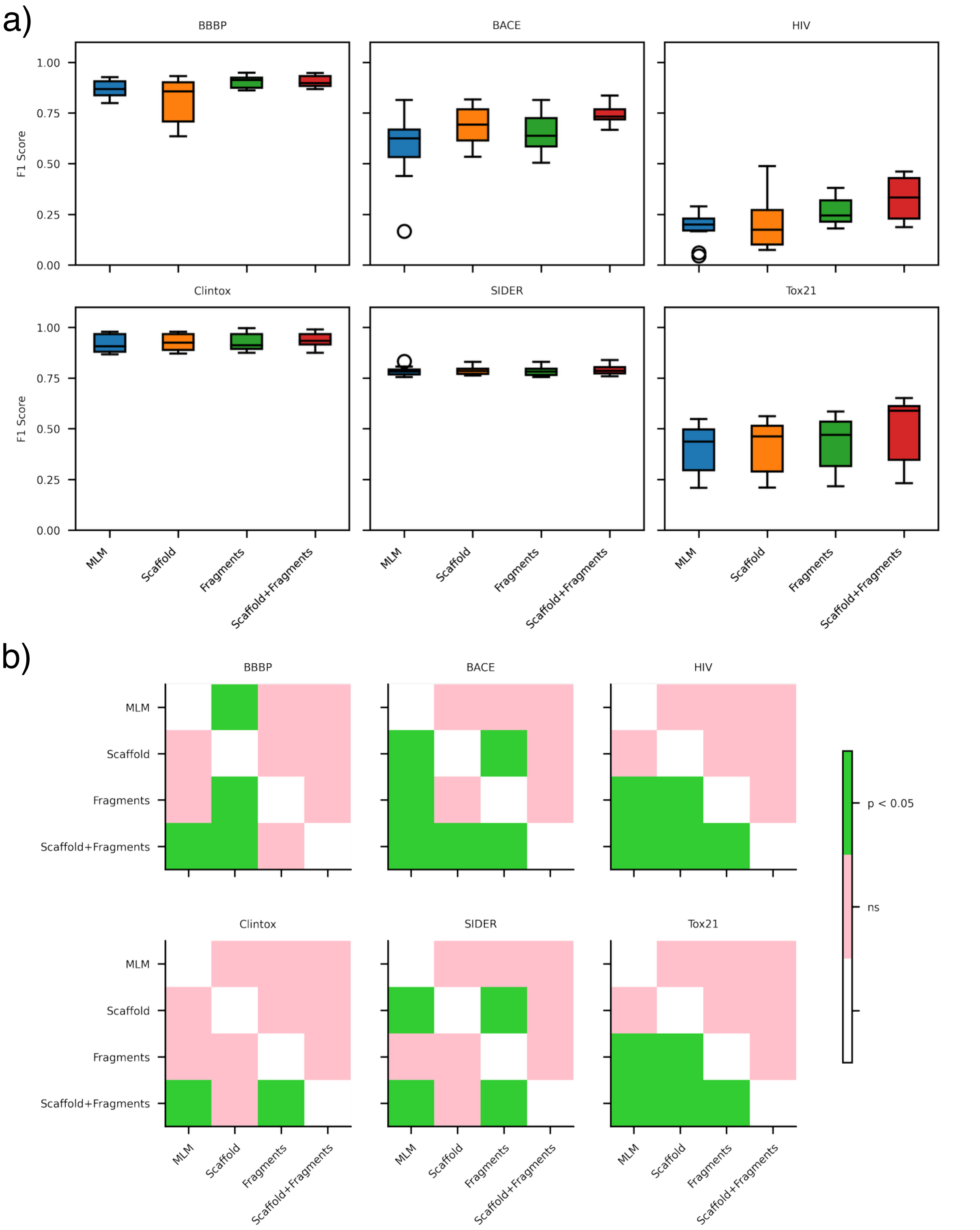

Figure 3: **Comparison of the downstream performance of four pretraining tasks.** a) Boxplot of the F1 scores from ten-fold splits using scaffold splitting and b) the statistical significance of the F1 scores using a Mann-Whitney U test and an alternative hypothesis that determines whether the distribution underlying the metrics from the model on the $y$ axis is stochastically greater than that of the model on the $x$ axis – green, p value is less than 0.05.

To test whether the pretraining tasks had any effect on the downstream performance, all the models were finetuned using six common classification benchmarks (see Table 2, Methods). Each benchmark dataset was split into training, validating, and testing datasets (with a ratio of 80/10/10) ten times using the scaffold splitting method and ten different random seeds.

Each model was then finetuned against each of the ten dataset splits for each benchmark. The F1 score, which is the harmonic mean between the precision and recall, was calculated using the predictions and ground truth of the test datasets. Typically the area under the curve of the receiver operator characteristic (ROC AUC) is used to compare molecular property prediction models, but some have found this metric to overestimate the model's performance against the test data [27]. To test the statistical significance of the results between the different models, the ten test scores for each benchmark were taken and p-values were calculated using a Mann-Whitney U test [28]. Deng et al., provide a useful discussion on the application of different statistical analyses in molecular property prediction, and the Mann-Whitney U test was chosen as a result of their findings [27]. The figures in this paper that report statistical significance show whether the model on the y-axis performs statistically greater than the model on the x-axis against the ten test datasets and are highlighted in green when $p < 0.05$.

Here, we find that the Scaffold+Fragments task statistically outperforms the MLM task in all six benchmarks, the standalone Scaffold task in four benchmarks and the Fragments task in five benchmarks. In only one case does the MLM task outperform the domain-specific tasks. From these comparisons, domain-specific pretraining tasks show clear improvements over the MLM task. Moreover, pretraining on multiple domain-specific tasks further improves downstream finetuning performance over pretraining on a single task. For this reason, all other models in this study were pretrained on the Scaffold+Fragments task and the zinc15 1M dataset, unless otherwise stated.

## 2.2 Comparison to Other Models

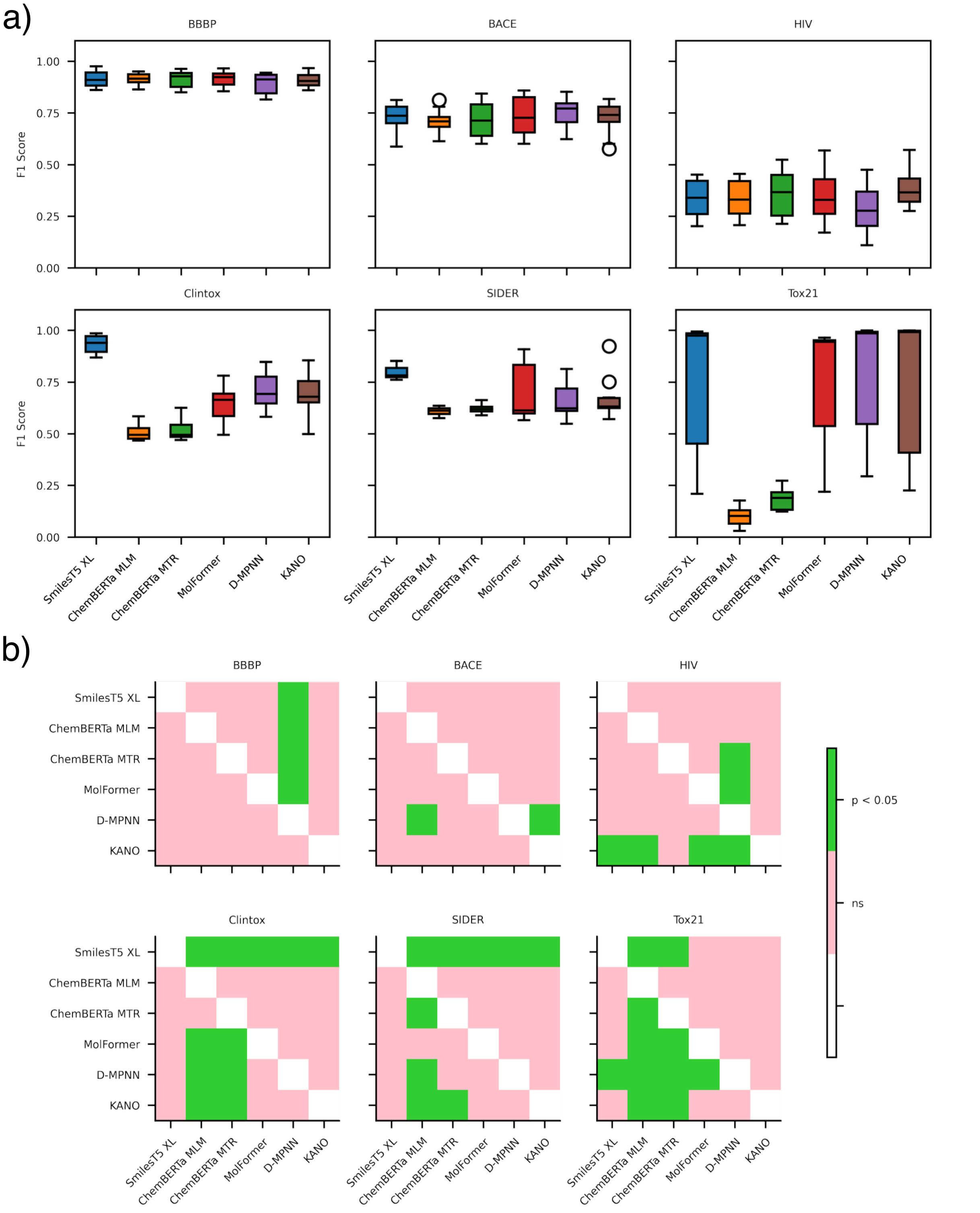

Figure 4: **Comparison of the downstream performance of the SmilesT5 XL model and a selection of other publicly available models.** a) Boxplot of the F1 scores from ten-fold splits using scaffold splitting and b) the statistical significance of the F1 scores using a Mann-Whitney U test and an alternative hypothesis that tests whether the distribution underlying the metrics from the model on the $y$ axis is stochastically greater than that of the model on the $x$ axis.

Here, we wanted to compare our model to best performing models that utilise either graph-based or language-based methods. For the language-based models the MolFormer model published by IBM researchers [29], and two ChemBERTa-2 models published by Reverie Labs [14, 15] were chosen. For comparison to graph-based models, the D-MPPN model (Yang et al.) [6] and the KANO model (Fang et al.) [10] were chosen. All the models were chosen as they are generally the best performing models across a range of benchmarks and offer a unique approach to molecular property prediction allowing us to compare our model on broad scale.

To this end, we pretrained an *xl* sized model against the zinc15 1m dataset (SmilesT5 XL) Here we find that the SmilesT5 XL model statistically outperforms all other language and graph models in both the Clintox and SIDER benchmarks. The SmilesT5 XL model is only bettered twice across all benchmarks, the first against the KANO model in the HIV benchmark, and the second against the D-MPNN model in the Tox21 benchmark. These are the two largest datasets of the benchmarks tested so the advantages associated with pretraining on a large corpus prior to finetuning are diminished in these datasets. SmilesT5 XL, however, outperforms the other models on thirteen occasions, highlighting the advantage to using domain-specific pretraining tasks and the impact they can have when finetuning on downstream tasks. The SmilesT5 XL model is not outperformed by any of the other language models in any of the benchmarks.

Through capturing important information about the structure and molecular properties of molecules during pretraining, the SmilesT5 XL model is able transfer domain-specific knowledge to downstream tasks, resulting in significantly improved performance, as demonstrated here.

## 2.3 Ablation Studies

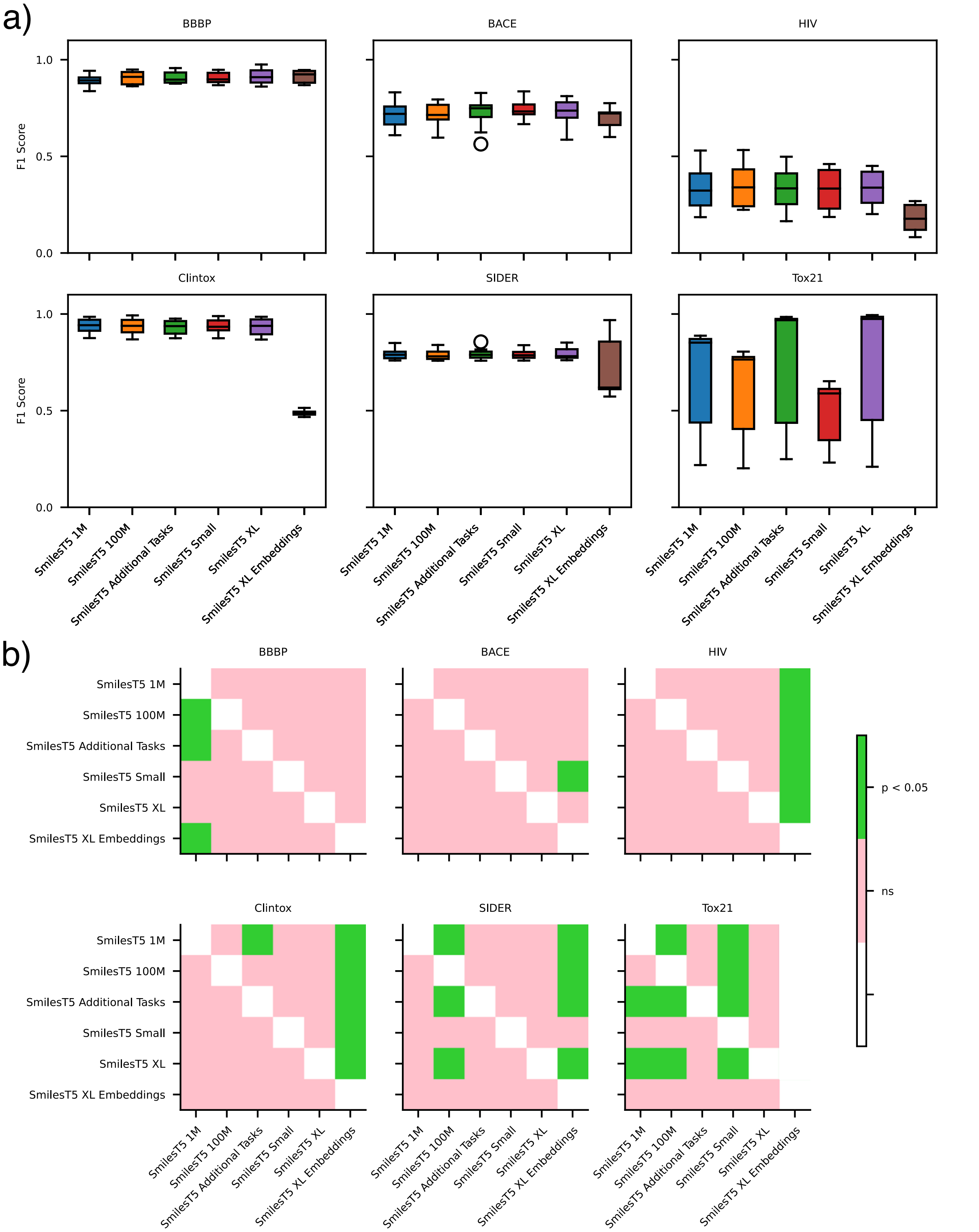

Figure 5: **Comparison of the downstream performance of multiple ablation studies.** a) Boxplot of the F1 scores from ten-fold splits using scaffold splitting and b) the statistical significance of the F1 scores using a Mann-Whitney U test and an alternative hypothesis that determines whether the distribution underlying the metrics from the model on the $y$ axis is stochastically greater than that of the model on the $x$ axis.

To further understand the inner workings of the SmilesT5 model, experiments were performed that tested how dataset size, parameter count, different finetuning tasks and sequence embeddings impacted the performance of the model against classification benchmarks.

**Dataset Size.** Large language models are typically pretrained on many billions of tokens. To explore whether this is necessary for pretraining on the molecular knowledge tasks outlined in this work, two *base*-sized models were pretrained on two different datasets. The first model was pretrained on 1 million molecules and the second was pretrained on 100 million molecules from the Zinc15 [30] dataset (labelled SmilesT5 1M and 100M, respectively, see Figure 5). Each model was then finetuned against the six benchmarks as outlined above.

Here it was found that increasing the pretraining dataset size from 1m to 100m molecules only improved the performance in one benchmark, BBBP. Intriguingly, in the SIDER and Tox21 benchmarks, the model pretrained with 1m molecules outperforms the model pretrained with 100m molecules. The bilingual evaluation understudy (BLEU) score and word rate error for the model pretrained on 100m molecules is within the range of the model pretrained with 1m molecules (see Table S1), suggesting that while the model sees more examples of unique molecules when pretrained with a larger dataset, it does not require such a large dataset to learn the underlying algorithm of the pretraining tasks. With the costs associated with pretraining large language models ever increasing, it is clear that with more appropriate, domain-specific pretraining, data efficiency can be improved, and molecular language models do not have to rely upon extremely large datasets to learn the underlying chemistry of molecules.

**Model Size.** Another method to improve language model performance is to increase the number of parameters in the model. To test this *small*, *base* and *XL*-sized models (see Table 1 for model configurations) were pretrained on 1m molecules using the Scaffold+Fragments joint pretraining task (see Figure 5, labelled SmilesT5 Small, SmilesT5 1M and SmilesT5 XL, respectively).

In only one benchmark, Tox21, does the *XL* model outperform the *small* and *base* models, with the *base* model also outperforming the *small* model in this benchmark. This suggests that in the context of this benchmark, there is a slight trend that model size matters. This does not, however, occur in the other five benchmarks, suggesting that the data and model efficiency provided by the domain-specific pretraining tasks allows for highly performing models to be trained with much lower computational overhead.

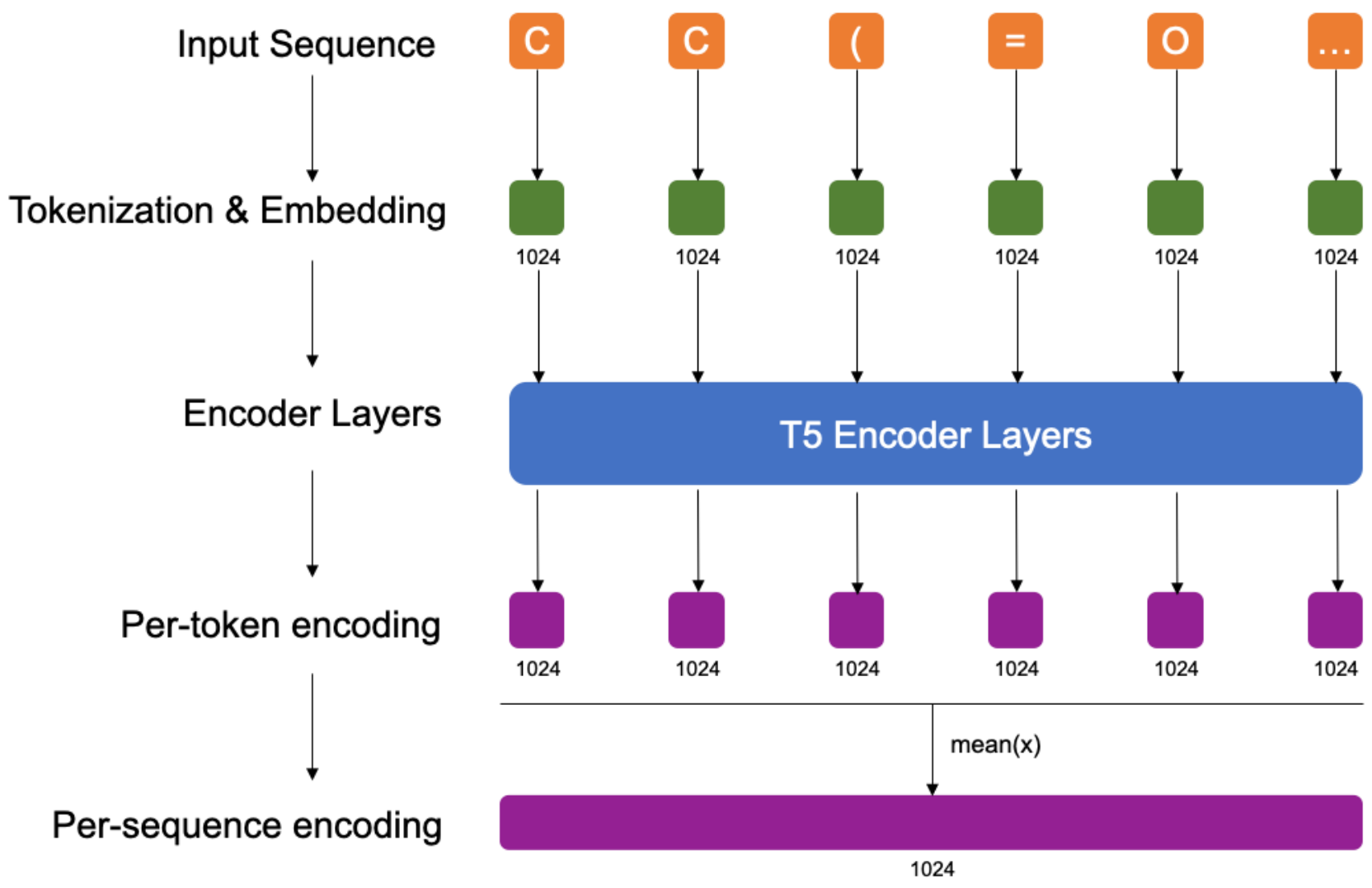

Figure 6: **Feature extraction method using the encoder layers of the SmilesT5 model.**

**Embeddings.** To further explore minimising computational cost, the pretrained model was used to obtain per-sequence embeddings that were then used as inputs for downstream tasks using a random forest classifier. As finetuning large language models typically requires access to expensive top tier hardware, which is not available to all, this could offer a much lower entry barrier for informaticians. Due to missing labels in the Tox21 dataset, this benchmark was omitted from this analysis.

In three of the five benchmarks (HIV, Clintox, and SIDER) finetuning the *XL* model statistically outperforms the random forest trained on the *XL* embeddings (Figure 5). While this clearly shows the advantage of finetuning the whole model, it is worth noting that the embeddings of *XL* model statistically outperform finetuning the *base* model in the BBBP benchmark, suggesting that these embeddings could be used if access to high performance computing is not available. It is also conceivable that these embeddings could be used as molecular fingerprints, as they encapsulate information about the scaffold and fragments present in a molecule and give a fixed-length vector representation.

**Additional Finetuning Tasks.** One advantage to the text-to-text approach while using a T5 model is that the model can be pretrained, and also finetuned, on multiple tasks simultaneously. To explore the impact of this on finetuning, the pretraining tasks were included during the finetuning process so that the model was finetuned on the downstream tasks (e.g. BBBP benchmark task) as well as the Scaffold and Fragments tasks. If the model can learn general molecular knowledge of the relevant molecules at the same time as the specific downstream task, the model may be able to better understand the underlying chemistry behind the downstream task. For this, the same task tokens were prepended as those used during pretraining, with an additional task token of "labels:" prepended to the SMILES strings for the downstream task.

Finetuning a *base*-sized model on the additional tasks (labelled SmilesT5 Additional Tasks, see Figure 5), improved the downstream performance over the *base* SmilesT5 1M model, with statistical significance ($p < 0.05$), in the BBBP and Tox21 benchmarks. This is countered by the model demonstrating a statistically lower performance in the Clintox benchmark. From these results, it is difficult to say whether finetuning on additional tasks improves overall finetuning performance. It is, however, worth noting that for every $n$ tasks added during finetuning the time it takes to finetune increases $n$-times, so choosing the correct finetuning tasks is important.

The flexibility of the text-to-text approach, however, allows for many additional tasks to be explored and finetuning with tasks other than the pretraining tasks may well improve performance further. Finetuning alongside a task that trains the model on the solubility of the molecules, for example, may well increase performance when learning certain datasets.

# 3 Conclusion

Effective pretraining tasks are crucial for LLMs and existing work that aims to learn the language of molecules fail to utilise address this. The text-to-text framework and the two novel domain-specific pretraining tasks presented here have been shown to significantly improve performance over existing graph- and language-based approaches in a series of molecular property classification benchmarks. Masked language modelling, which is typically used for learning the language of molecules, has been shown to lead to poorer downstream performance for molecular property prediction. The flexibility of this approach not only allows the SmilesT5 models to transfer domain-specific knowledge captured during pretraining to specific downstream tasks, but it also opens up the possibility to both pretrain and finetune on many tasks simultaneously.

Through a number of experiments the impact of both dataset size and model parameter count on finetuning performance was explored. It was found that increasing these only led to marginal improvements during finetuning, demonstrating how even a small model using a small dataset is able to capture important molecular information. The training of random forest classifiers on the embeddings from the pretrained SmilesT5 model was also explored and offers the opportunity for informaticians to utilise the molecular information captured by the SmilesT5 model without needing access to high performance computing.

# 4 Methods

## 4.1 Pretraining.

All models pretrained models were trained using 4x NVIDIA A100 graphics processing units.

**Masked Language Modelling.** The masked language modelling proposed in the original T5 paper [21] was used with 15% of the input sequence masked.

**Scaffold.** The Murcko scaffold SMILES strings for each molecule in the training set was generated and then the model was trained to reconstruct the scaffold SMILES from the input molecular SMILES (see Figure 1b).

**Fragments.** RDKit was used to generate the presence of 86 fragments (see Table S2) for each molecule with the target text for each molecule being a sequence of tokens that represent the names of the fragments that are present in the molecule, with the input text as the molecular SMILES (see Figure 1c).

**Scaffold + Fragments.** A token was prepended to the SMILES string inputs to indicate to the model which task was being performed. The tokens added were *scaffold:* and *fragments:* for the scaffold and fragments tasks, respectively (see Figure 1d). For this task, the number of training steps used was set to be equal to those used in the individual tasks.

**Model Parameters.** The model sizes used in this study were derived from the *base* and *XL* models defined in the original T5 paper, with an additional *small* model being created (see Table 1) [21].

**Table 1 - Details of the model parameter sizes of the models**

| | Small | Base | XL |
|---|---|---|---|
| Embedding size | 512 | 768 | 1 024 |
| Number of layers | 6 | 12 | 24 |
| Number of heads | 8 | 12 | 32 |
| Intermediate feed forward size | 2 048 | 3 072 | 16 384 |
| Total Parameters (M) | 4 | 198 | 2 215 |

**Pretraining Datasets.** The datasets used for pretraining were the 1 million and 100 million Zinc15 datasets hosted by DeepChem [26].

## 4.2 Finetuning

All models were finetuned using a single NVIDIA A100 graphics processing unit.

**Sequence Classification Model.** For finetuning the T5-based models on the SMILES strings of the selected benchmarks, the tasks were first converted to text-to-text tasks. For each molecule the labels were converted into the target sequence text so that if a molecule has a certain label, then the target sequence would contain that label or be an empty string if it does not have a label. For multilabel tasks, such as SIDER or Tox21, the target sequence would be the sequence of the labels that correspond to the molecule, or an empty string if no labels correspond to the molecule.

**Benchmark Datasets.** The benchmark datasets (see Table 2) were obtained from DeepChem [26] and were split using the scaffold splitting method outlined in MoleculeNet [2]. The data were split into train/valid/test with ratios of 80/10/10 using ten different seeds. All models, including publicly available models, were then finetuned on the same ten splits of each benchmark dataset.

**Table 2 - Details of each dataset, including the name, number of molecules, number of classes and the evaluation metric**

| Dataset | Number of Molecules | Number of Tasks |
|---|---|---|
| BBBP | 2 053 | 1 |
| BACE | 1 522 | 1 |
| HIV | 41 913 | 1 |
| Clintox | 1 491 | 2 |
| SIDER | 1 427 | 27 |
| Tox21 | 8 014 | 12 |

**Metrics Calculations.** For all tasks, each unique label in the dataset were added to the tokenizer and the model embedding layers. The outputs from a model's language head were taken (with a shape of [batch size, sequence length, vocabulary size]) and a softmax operation was performed for each token across the vocabulary. The maximum was then taken across the output sequence, resulting a per-molecule vector that represents the maximum probability of the model outputting each token in the vocabulary (with a shape of [batch size, vocabulary size]). The probabilities for only the label tokens were then taken resulting in a final vector of the probabilities for the model outputting each label (with a shape of [batch size, number of labels]). For single label classification tasks, e.g. the BBBP benchmark, the final vector length is 1. These probabilities were used to calculate the metrics against the validation and test datasets.

In some of our results it was found that the ROC AUC was over-optimistic in the performance of the models, particularly the ChemBERTa models against the Tox21 dataset, where the ROC AUC scores are *ca* 0.70 and the F1 scores are *ca* 0.20 (see Figures S8 and S9). This highlights the importance of using the correct metrics when drawing conclusions upon the performance of the property prediction models. Figures S2-11 show the ROC AUC and PR AUC of each experiment.

The F1 score, ROC AUC and PRAUC were calculated using the *torcheval* python package [31], the code for which can be seen in the GitHub repository (https://github.com/hothousetx/smiles_t5). The Mann-Whitney U rank tests were performed using the *scipy* python package [32] with the alternative parameter set to *greater*.

## 4.3 Embeddings.

The sequence embeddings were obtained by passing the SMILES strings of each molecule through the encoder portion of the model. The per token embeddings of the final hidden layer of the encoder were taken and averaged over the tokens (excluding special tokens such as the padding token) to obtain a one-dimensional vector with the same length as the dimensions of the model embeddings (see Figure 6). These embeddings were then used to train a random forest classifier, using 256 estimators. The same datasets and splits were used for training the random forest as were used to finetune the language models. The random forest classifiers were trained using 4 cores of an Intel Xeon Gold 6326 CPU.

## Code availability

The source code of the work, as well as scripts for finetuning and inference of the SmilesT5 model are available in the GitHub repository (https://github.com/hothousetx/smiles_t5). The SmilesT5 model is also available for downloading in the HuggingFace repository (https://huggingface.co/hothousetx/smiles_t5).

## Acknowledgements

This work was supported by Alan Turing Award 'Towards machine learning-driven prediction of the product chemical space of oxidosqualene cyclases'. AO also acknowledges funding from the Biotechnological and Biological Sciences Research Council (BBSRC) Institute Strategic Programme Grant 'Harnessing Biosynthesis for Sustainable Food and Health (HBio) (grant no. BB/X01097X/1), the John Innes Foundation, the Novozymes Prize 2023 (Novo Nordisk Foundation), and Wellcome Discovery Award #227375/Z/23/Z. We thank the Norwich Bioscience Institutes (NBI) Research Computing for computational support.

## Conflicts of Interest

P.S. and A.O are co-founders of HotHouse Therapeutics.

# Appendix

## S1 Pretraining Metrics

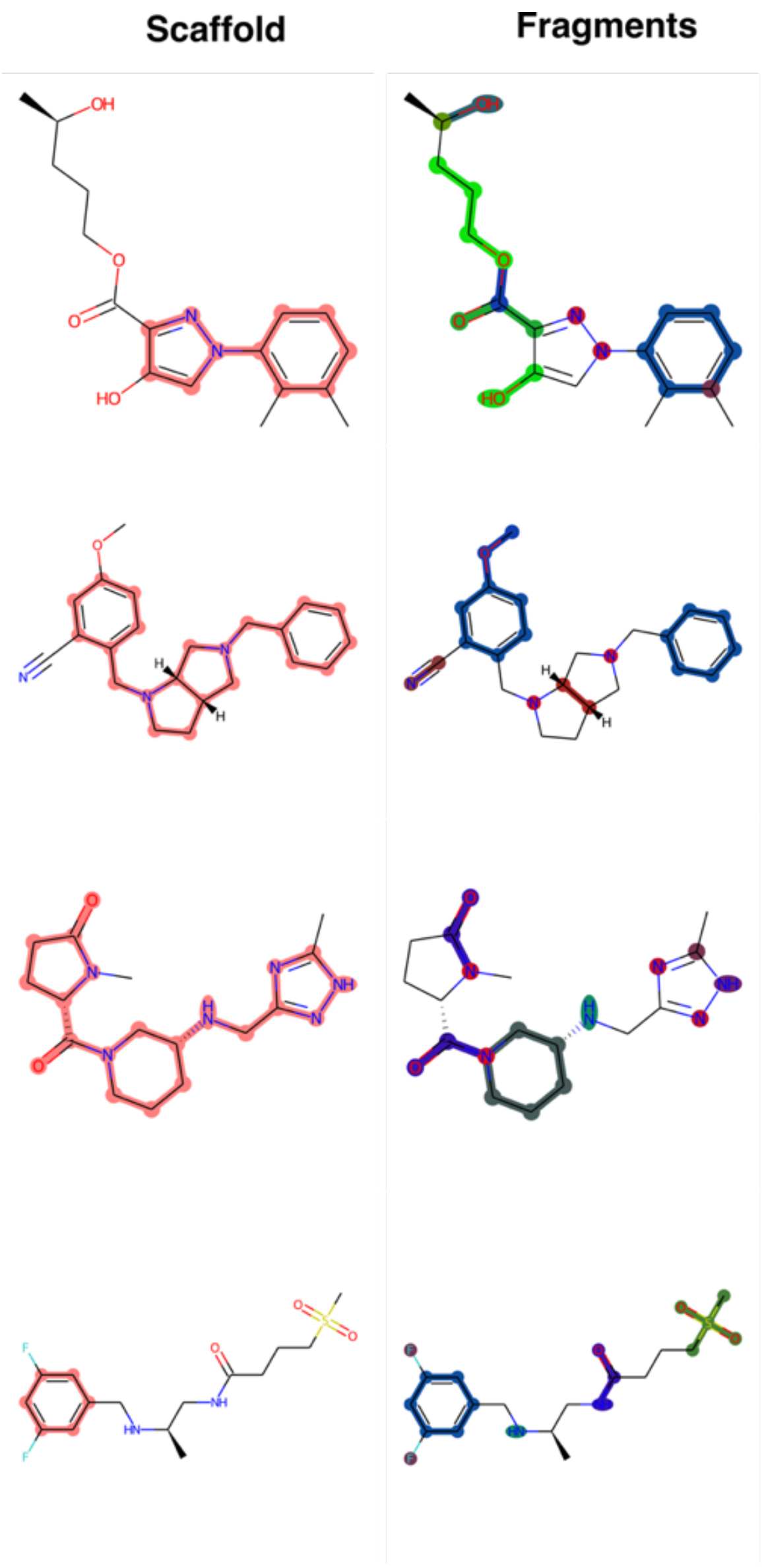

Figure S7: **Four randomly chosen examples from the pretraining test set of SmilesT5.** The first column shows the two-dimensional drawing of the molecule with the scaffold highlighted in red. The second column shows the same drawing of the molecule with the different fragments present highlighted in various colours. The third column shows the SMILES string used as the input, the scaffold SMILES string outputted by the model and the sequence of fragment tokens outputted by the model, with each fragment coloured the same as those in the second column, all model outputs are accurate to the target text.

Table S3: **The pretraining metrics for the SmilesT5 models pretrained on different dataset sizes.**

| Model | BLEU Score | Word Rate Error |
|---|---|---|
| SmilesT5 1M | 0.9967 | 0.004 |
| SmilesT5 100M | 0.9999 | 0.000 |

# S2 Finetuning Results

## S2.1 Task Comparisons

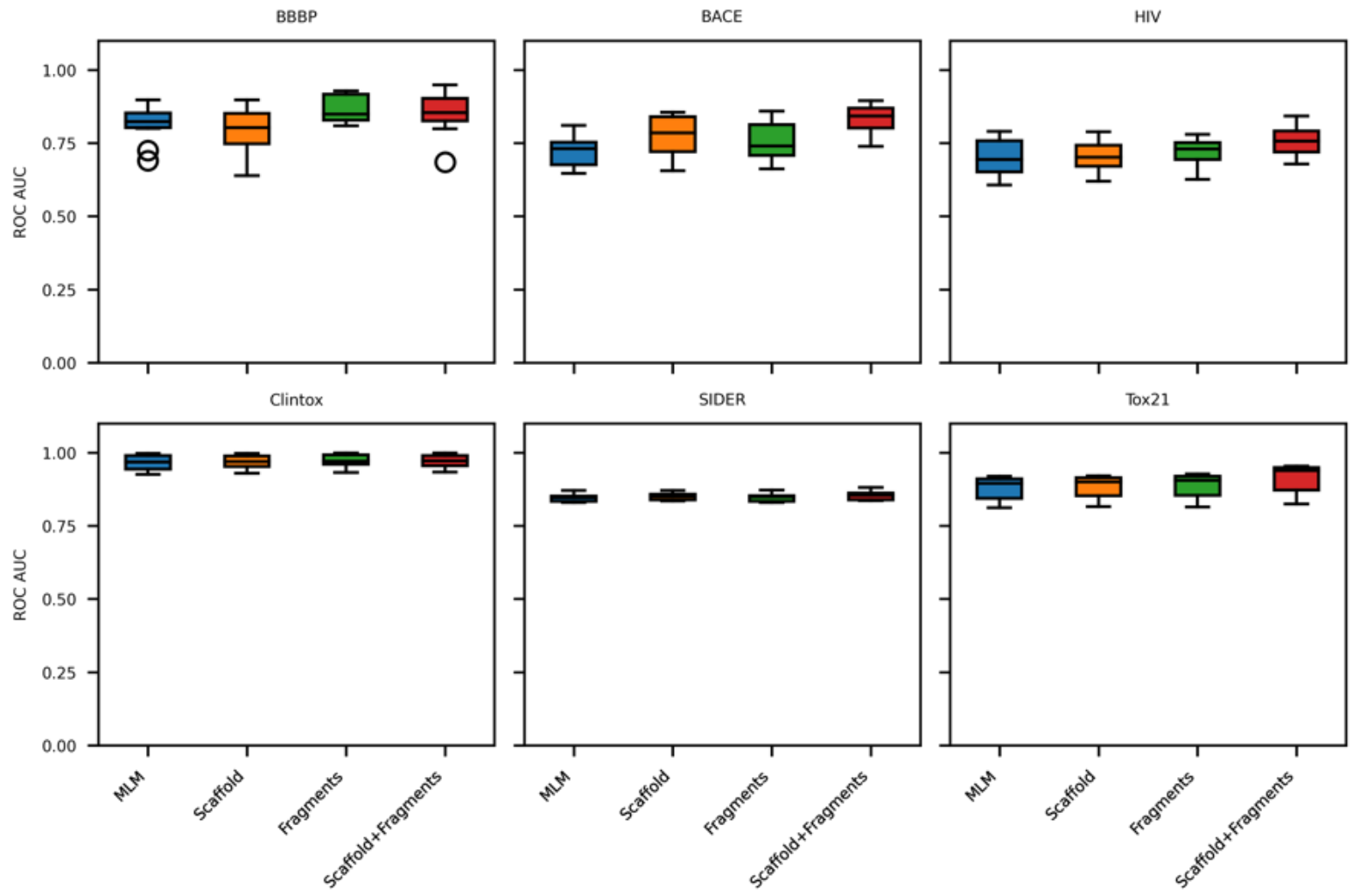

Figure S8: **Comparison of the downstream performance of four pretraining tasks.** Boxplot of the ROC AUC scores from ten-fold splits using scaffold splitting.

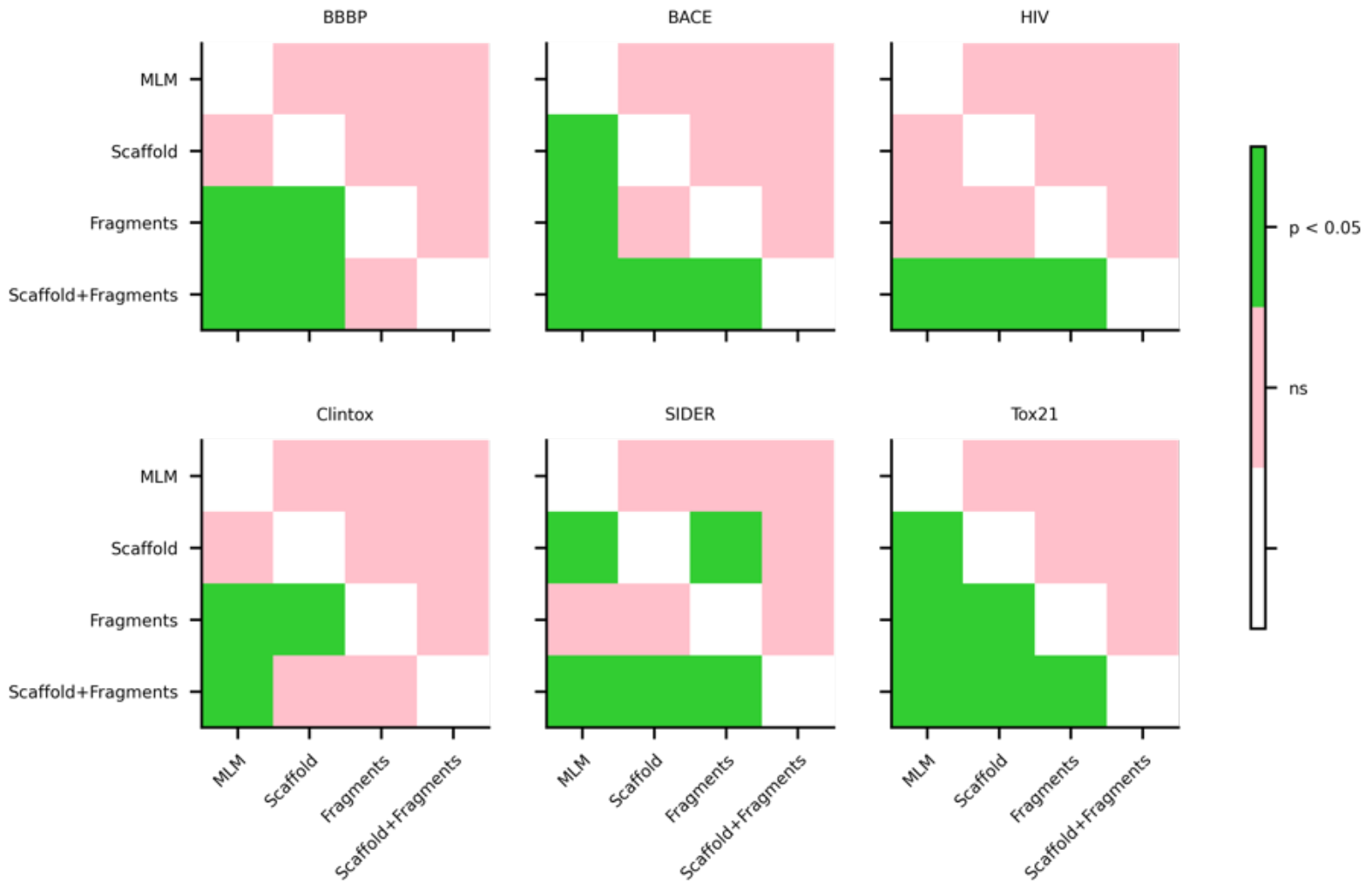

Figure S9: **Comparison of the downstream performance of four pretraining tasks.** The grid shows the statistical significance of the ROC AUC scores using the alternative hypothesis that determines whether the distribution underlying the metrics from the model on the $y$ axis is stochastically greater than that of the model on the $x$ axis, with green identifying occasions where the p value is less than 0.05.

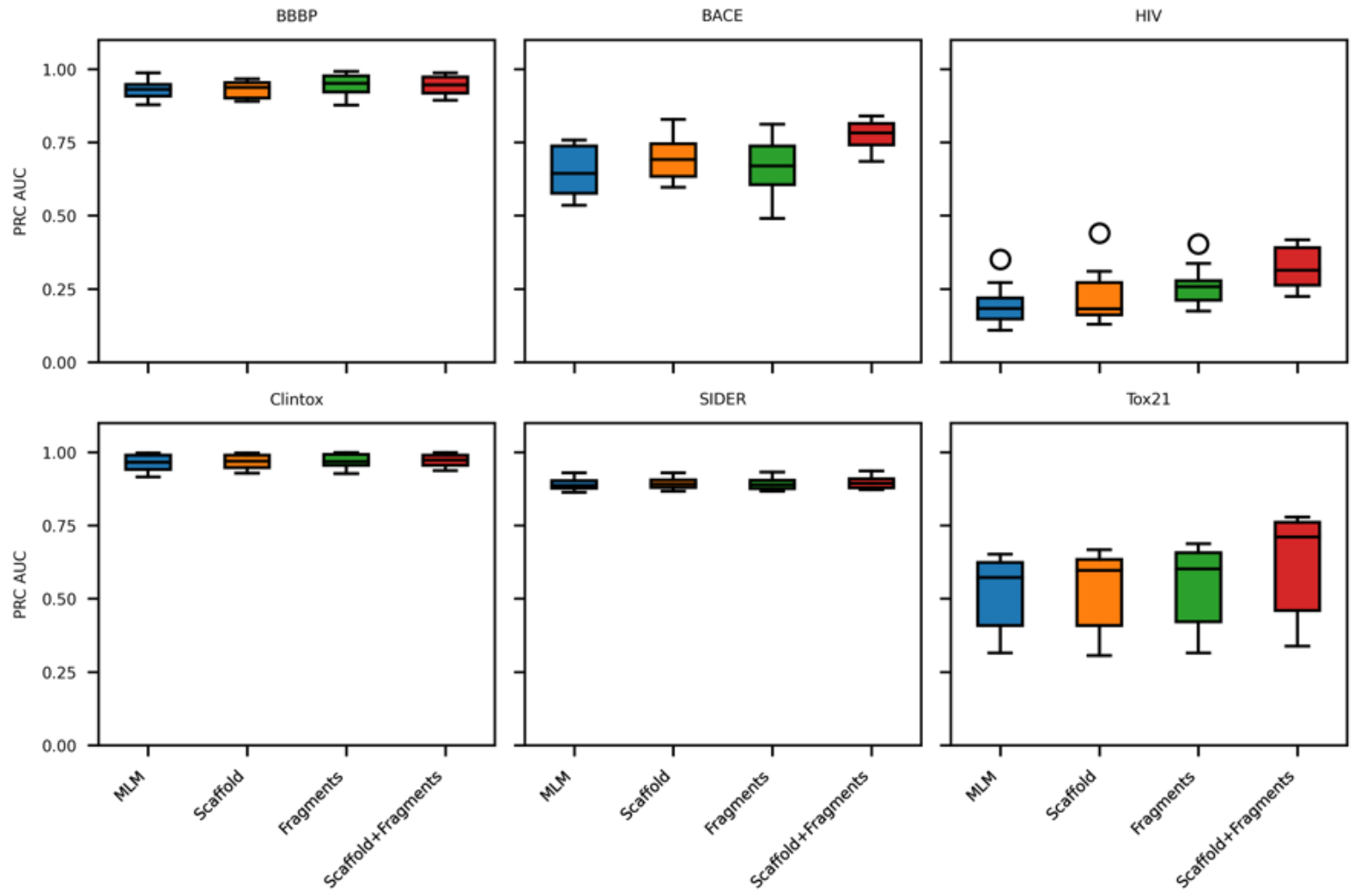

Figure S10: **Comparison of the downstream performance of four pretraining tasks.** Boxplot of the PR AUC scores from ten-fold splits using scaffold splitting.

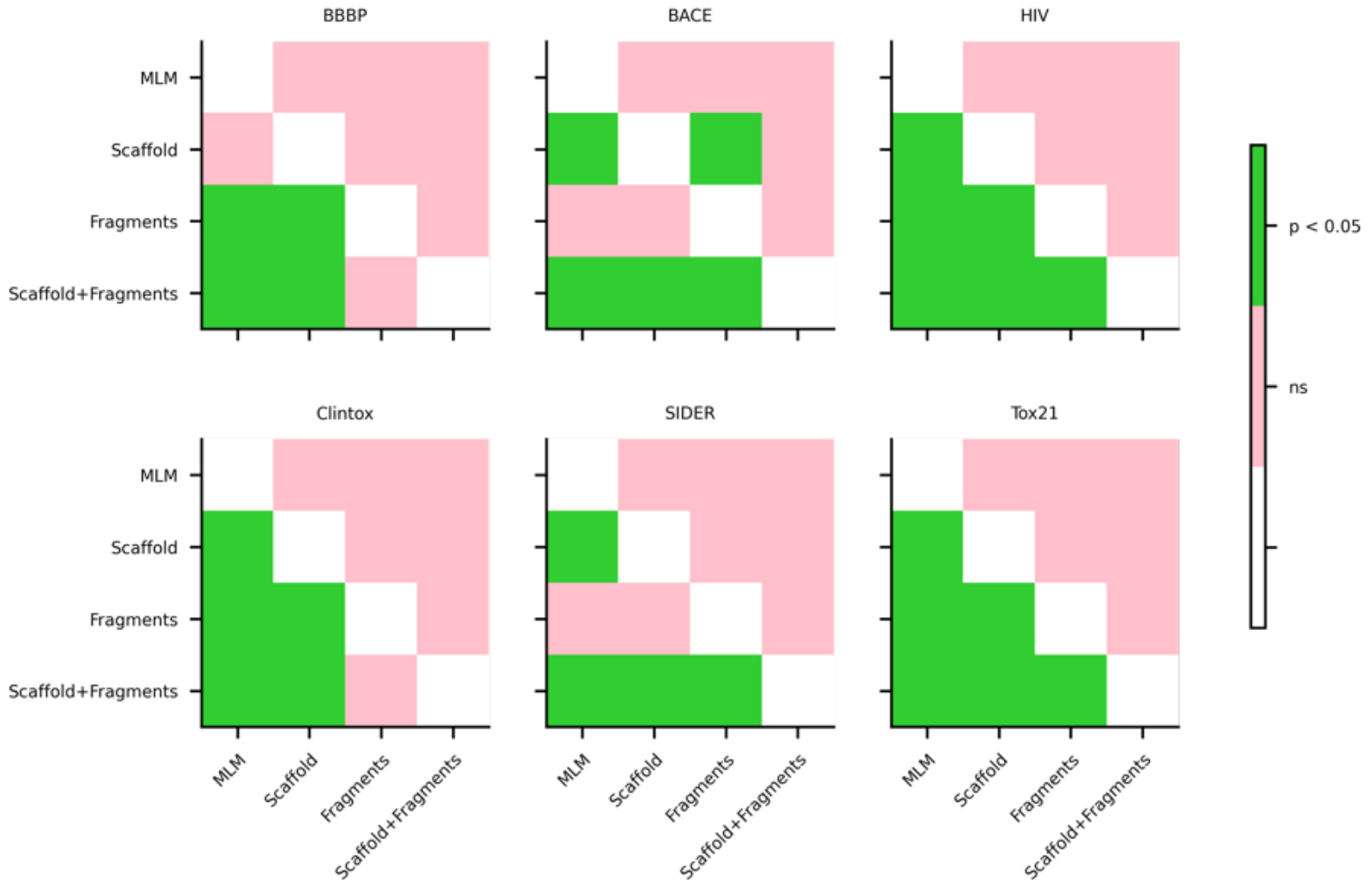

Figure S11: **Comparison of the downstream performance of four pretraining tasks.** The grid shows the statistical significance of the PRC AUC scores using the alternative hypothesis that determines whether the distribution underlying the metrics from the model on the $y$ axis is stochastically greater than that of the model on the $x$ axis, with green identifying occasions where the p value is less than 0.05.

## S2.2 Ablation Studies

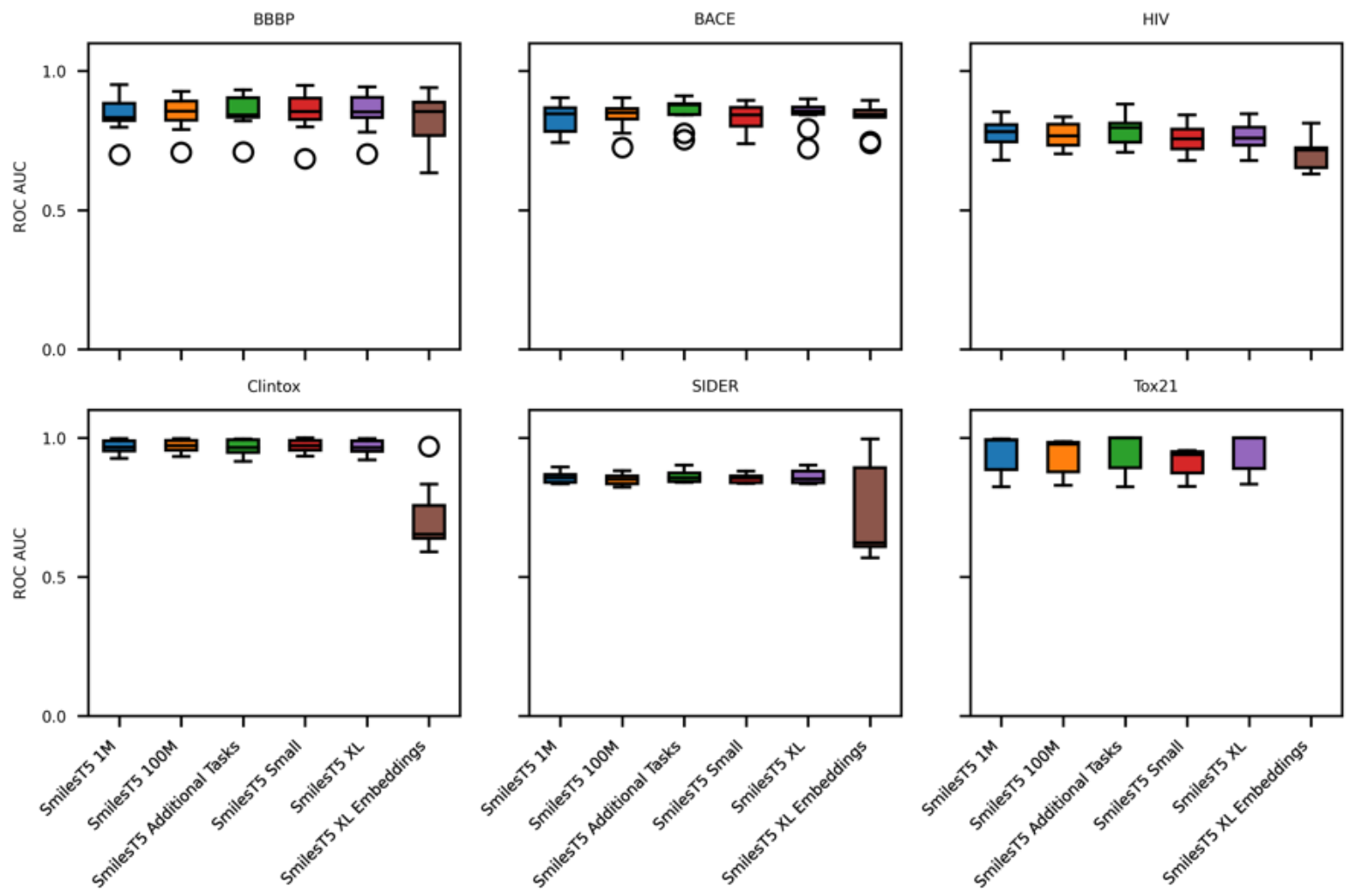

Figure S12: **Comparison of the downstream performance of multiple ablation studies.** Boxplot of the ROC AUC scores from ten-fold splits using scaffold splitting.

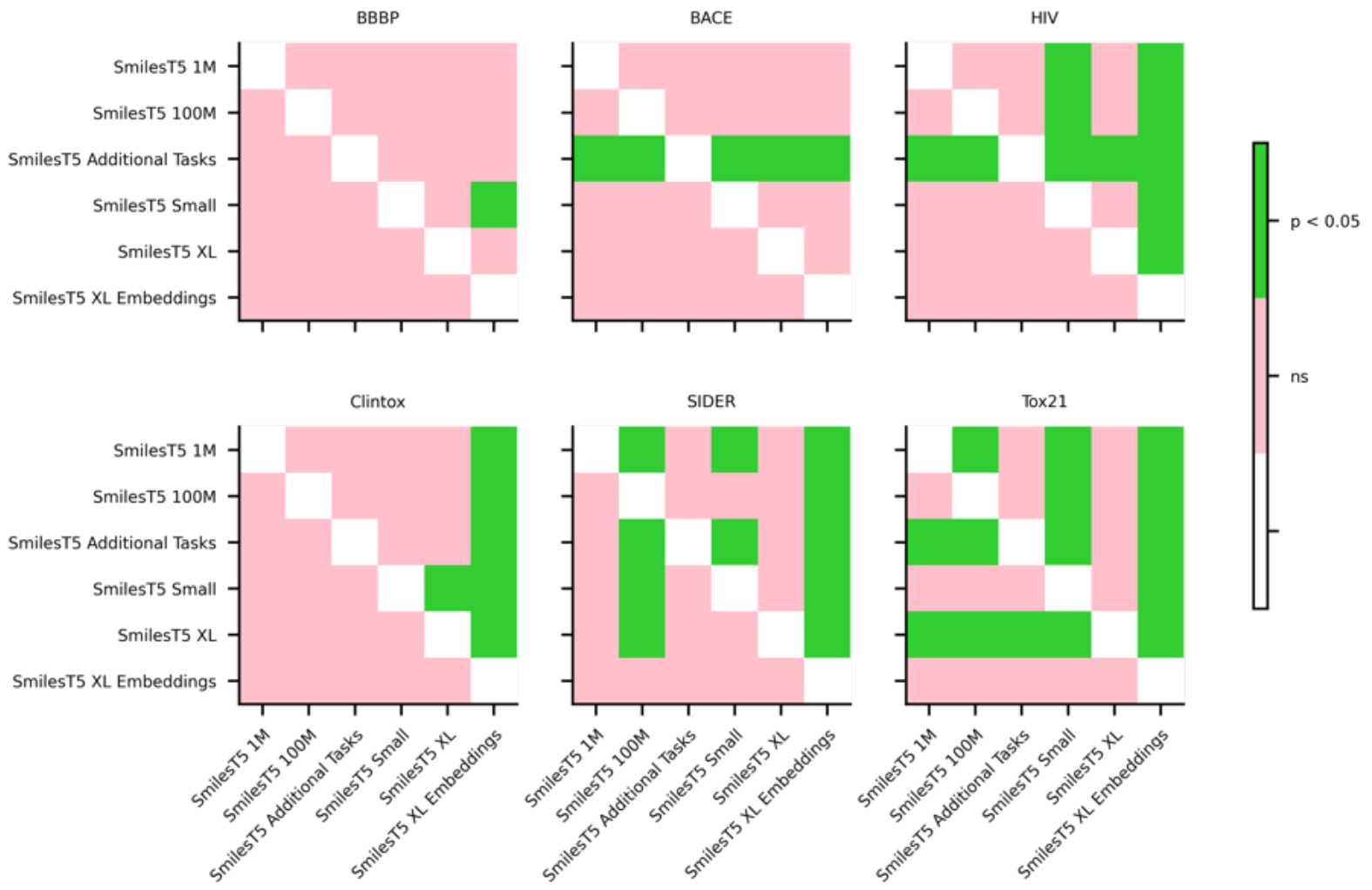

Figure S13: **Comparison of the downstream performance of multiple ablation studies.** The grid shows the statistical significance of the ROC AUC scores using the alternative hypothesis that determines whether the distribution underlying the metrics from the model on the $y$ axis is stochastically greater than that of the model on the $x$ axis.

## S2.3 Comparison to Other Models

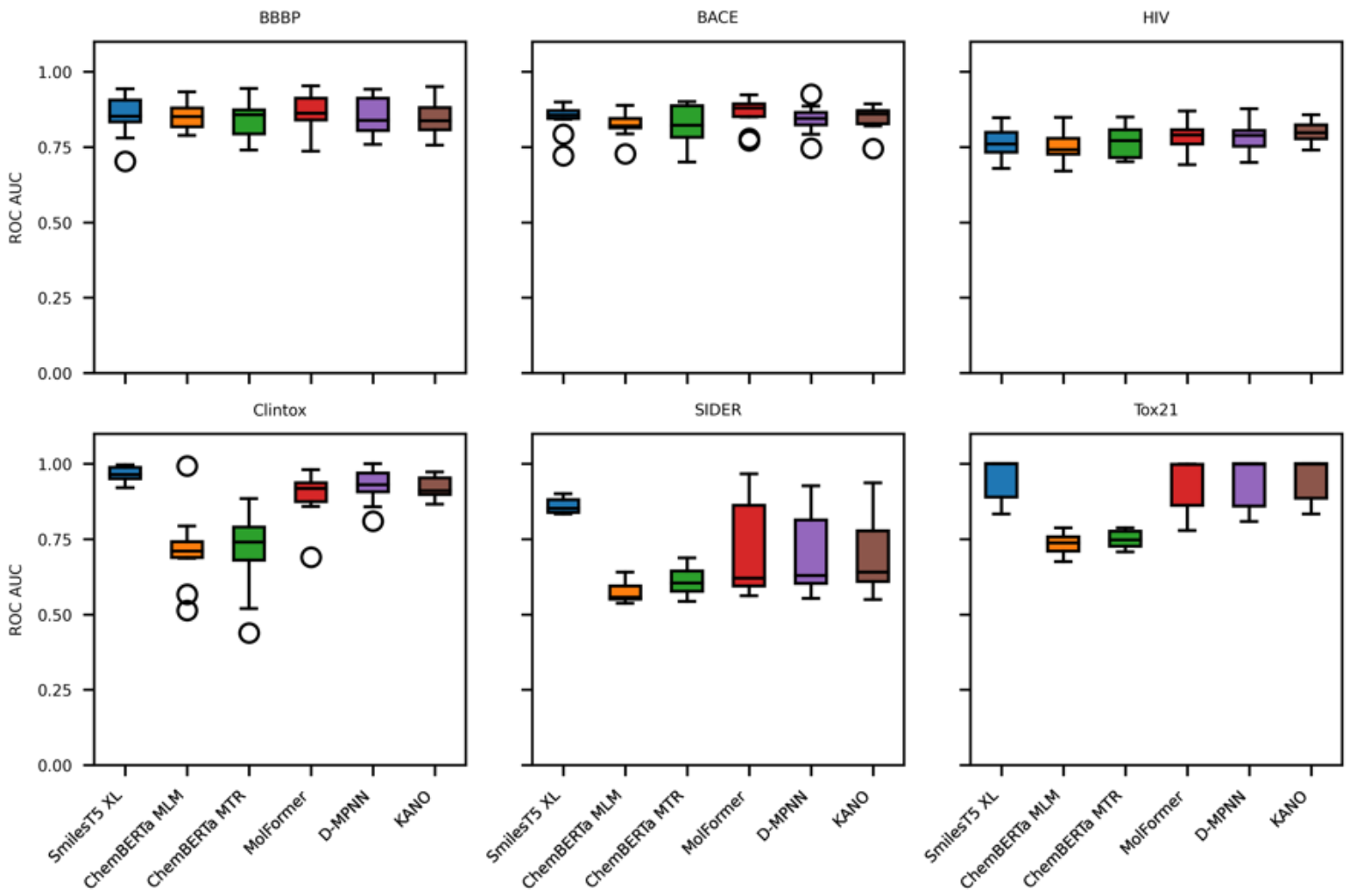

Figure S14: **Comparison of the downstream performance of the SmilesT5 XL model and a selection of other publicly available models.** Boxplot of the ROC AUC scores from ten-fold splits using scaffold splitting.

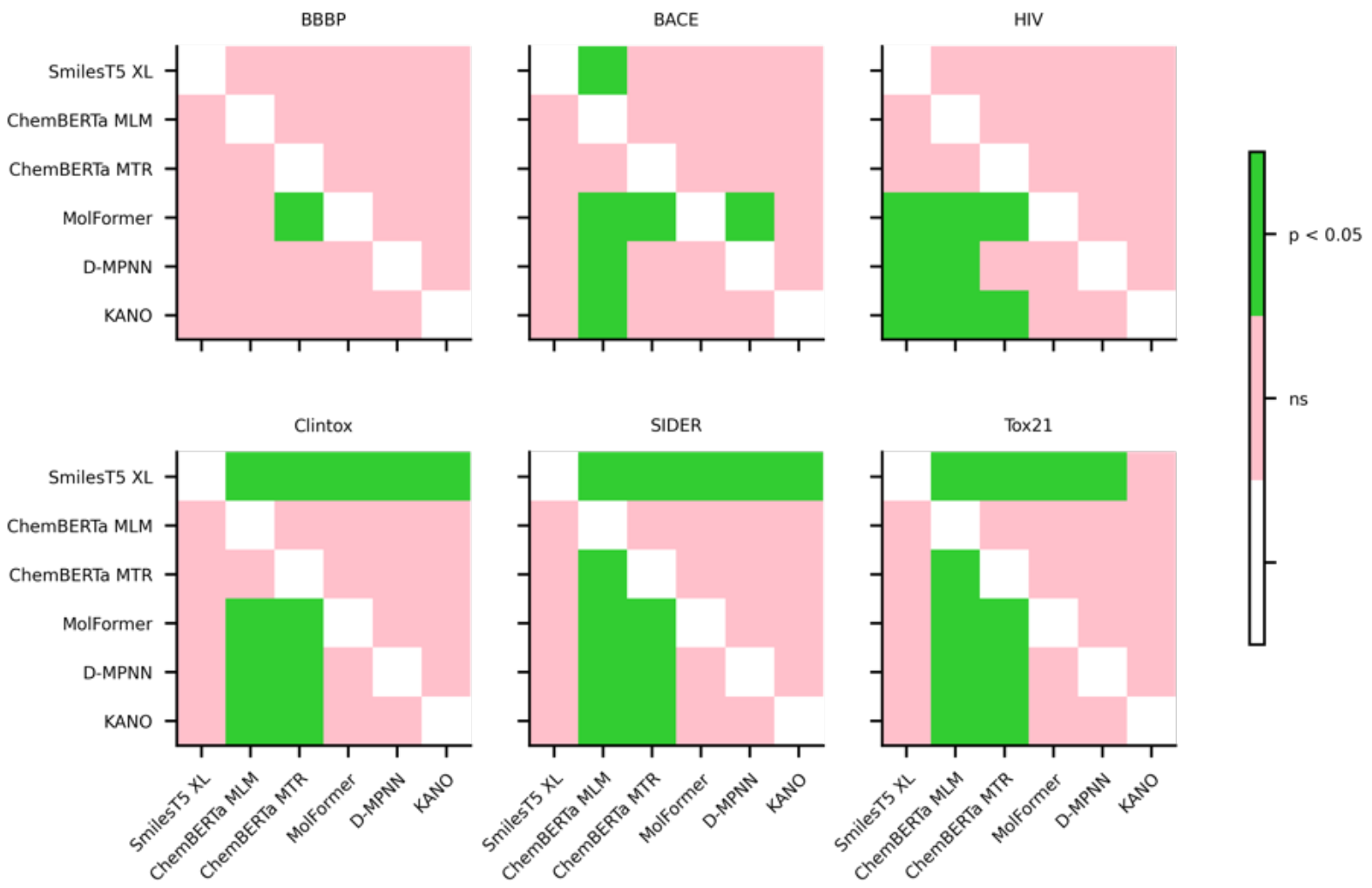

Figure S15 **Comparison of the downstream performance of the SmilesT5 XL and a selection of other publicly available models.** The grid shows the statistical significance of the ROC AUC scores using the alternative hypothesis that tests whether the distribution underlying the metrics from the model on the $y$ axis is stochastically greater than that of the model on the $x$ axis.

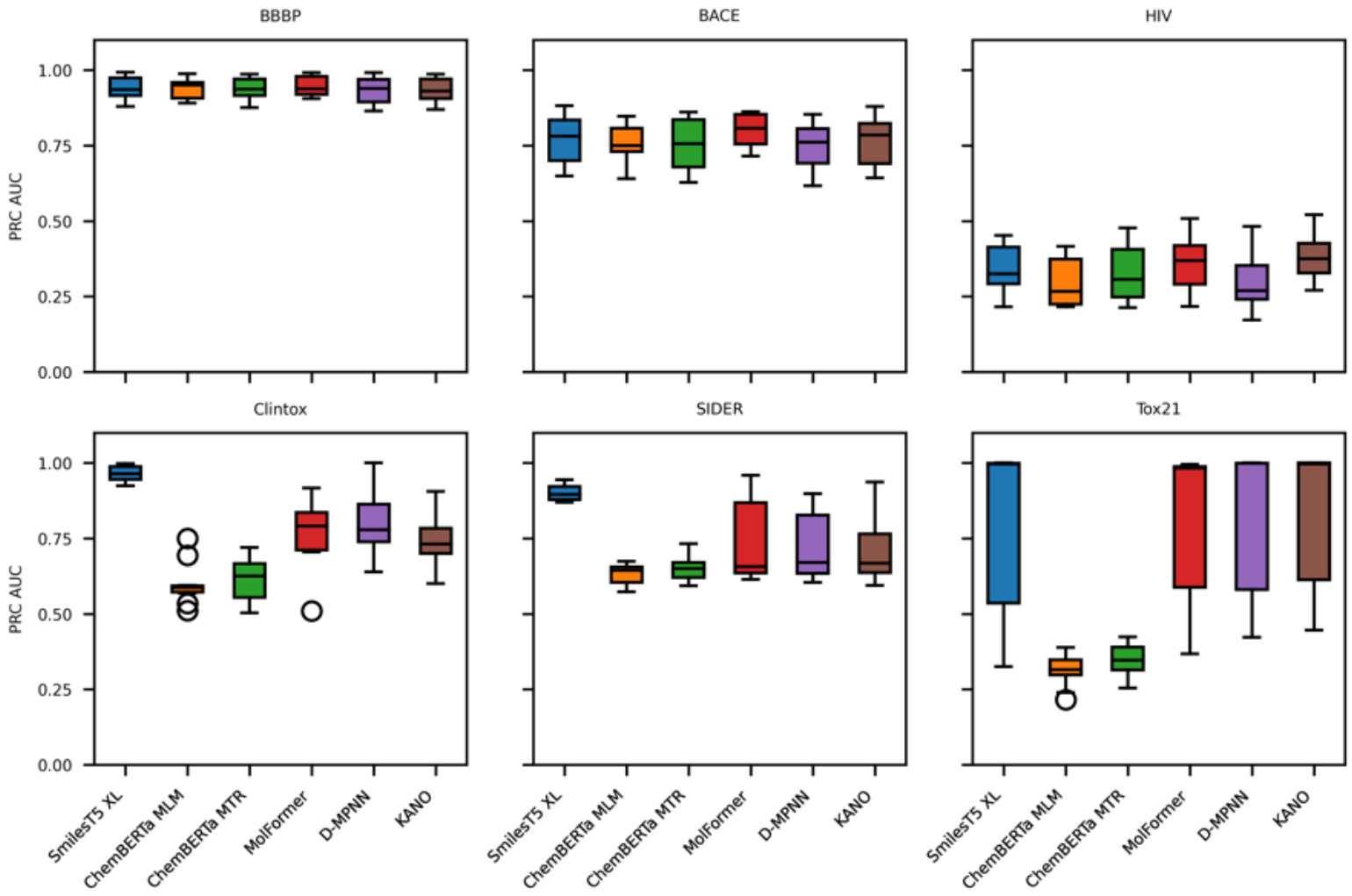

Figure S16: **Comparison of the downstream performance of the SmilesT5 XL and a selection of other publicly available models.** Boxplot of the PRC AUC scores from ten-fold splits using scaffold splitting.

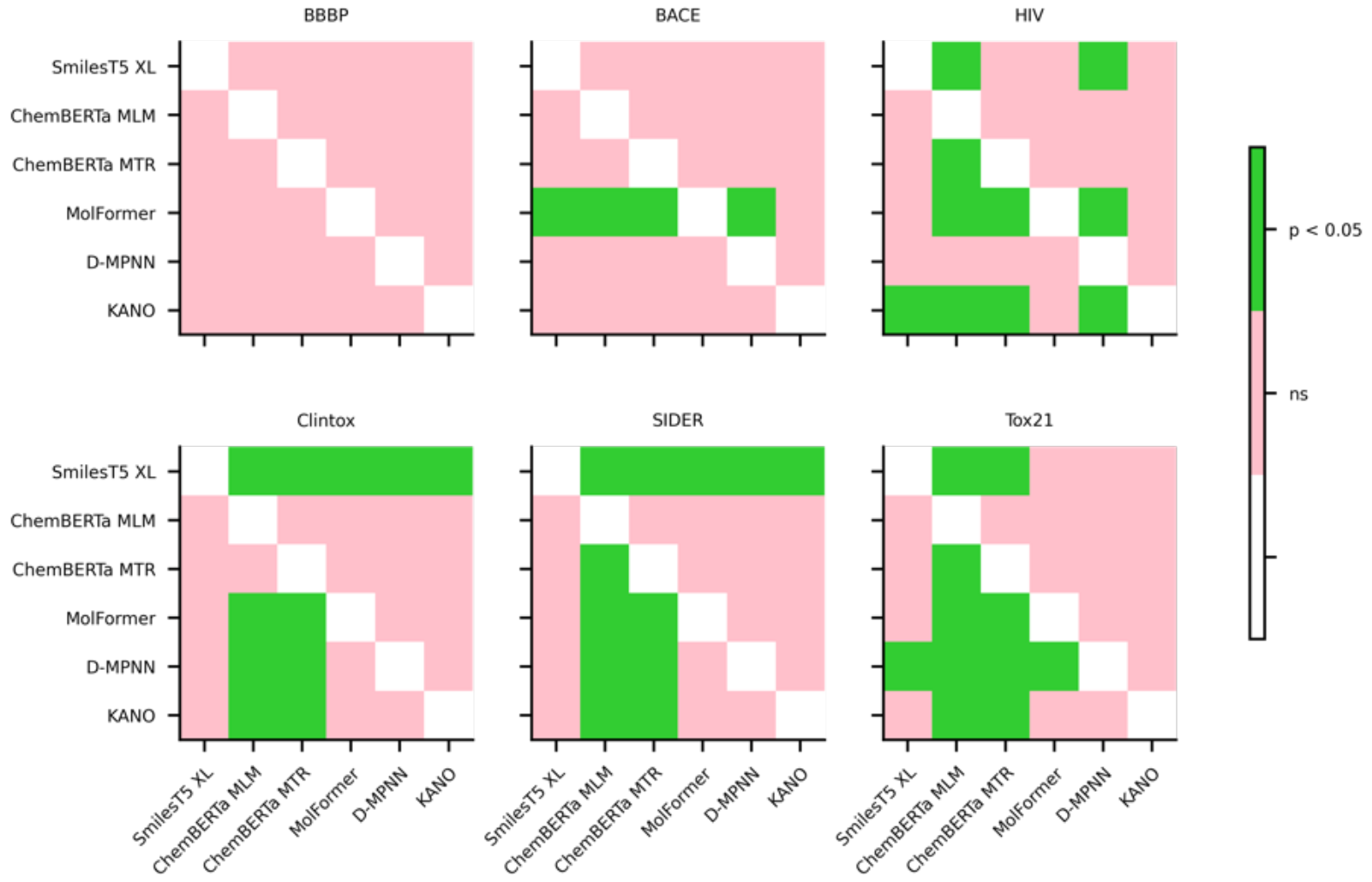

Figure S17 **Comparison of the downstream performance of the SmilesT5 XL and a selection of other publicly available models.** The grid shows the statistical significance of the PRC AUC scores using the alternative hypothesis that tests whether the distribution underlying the metrics from the model on the $y$ axis is stochastically greater than that of the model on the $x$ axis.

# S3 Methods

Table S4: **Token names, SMARTS strings, and brief descriptions of the fragments used during pretraining.**

| Token | Description | SMARTS |
|---|---|---|
| fr_C_O | Number of carbonyl O | [C&X3]=[O&X1] |
| fr_C_O_noCOO | Number of carbonyl O, excluding COOH | [C&!$(C-[O&H1])]=O |
| fr_Al_OH | Number of aliphatic hydroxyl groups | [C&!$(C=O)]-[O&H1] |
| fr_Ar_OH | Number of aromatic hydroxyl groups | c[O&H1] |
| fr_methoxy | Number of methoxy groups -OCH3 | [O&X2](-[#6])-[C&H3] |
| fr_oxime | Number of oxime groups | [C&X3]=[N&X2]-[O&X2] |
| fr_ester | Number of esters | [#6][C&X3](=O)[O&X2&H0][#6] |
| fr_Al_COO | Number of aliphatic carboxylic acids | C-C(=O)[O;H1,-] |
| fr_Ar_COO | Number of Aromatic carboxylic acide | c-C(=O)[O;H1,-] |
| fr_COO | Number of carboxylic acids | [#6]C(=O)[O;H1,-] |
| fr_COO2 | Number of carboxylic acids | [C&X3](=O)[O&X1&H0&-,O&X2&H1] |
| fr_ketone | Number of ketones | [#6][C&X3](=O)[#6] |
| fr_ether | Number of ether oxygens (including phenoxy) | [O&D2]([#6])[#6] |
| fr_phenol | Number of phenols | [O&X2&H1]-c1ccccc1 |
| fr_aldehyde | Number of aldehydes | [C&X3&H1](=O)[#6] |
| fr_quatN | Number of quarternary nitrogens | [$([N&X4&+]),$([N&X4]=*)] |
| fr_NH2 | Number of Primary amines | [N&H2,n&H2] |
| fr_NH1 | Number of Secondary amines | [N&H1,n&H1] |
| fr_NH0 | Number of Tertiary amines | [N&H0,n&H0] |
| fr_Ar_N | Number of aromatic nitrogens | n |
| fr_Ar_NH | Number of aromatic amines | [n&H1] |
| fr_aniline | Number of anilines | c-[N&X3&!$(N=*)] |
| fr_Imine | Number of Imines | [N&v3](=C)-[#6] |
| fr_nitrile | Number of nitriles | [N&X1]#[C&X2] |
| fr_hdrzine | Number of hydrazine groups | [N&X3]-[N&X3] |
| fr_hdrzone | Number of hydrazone groups | C=N-[N&X3] |
| fr_nitroso | Number of nitroso groups, excluding NO2 | [N&!$(N-O)]=O |
| fr_N_O | Number of hydroxylamine groups | [N&!$(N=O)](-O)-C |
| fr_nitro | Number of nitro groups | [$([N&X3](=O)=O),$([N&X3&+](=O)[O&-])][!#8] |
| fr_azo | Number of azo groups | [#6]-N=N-[#6] |
| fr_diazo | Number of diazo groups | [N&+]#N |
| fr_azide | Number of azide groups | [$(*-[N&X2&-]-[N&X2&+]#[N&X1]),$(*-[N&X2]=[N&X2&+]=[N&X1&-])] |
| fr_amide | Number of amides | C(=O)-N |
| fr_priamide | Number of primary amides | C(=O)-[N&H2] |
| fr_amidine | Number of amidine groups | C(=N)(-N)-[!#7] |
| fr_guanido | Number of guanidine groups | C(=N)(N)N |
| fr_Nhpyrrole | Number of H-pyrrole nitrogens | [n&H1] |
| fr_imide | Number of imide groups | N(-C=O)-C=O |
| fr_isocyan | Number of isocyanates | N=C=O |
| fr_isothiocyan | Number of isothiocyanates | N=C=S |
| fr_thiocyan | Number of thiocyanates | S-C#N |
| fr_halogen | Number of halogens | [#9,#17,#35,#53] |
| fr_alkyl_halide | Number of alkyl halides | [C&X4]-[Cl,Br,I,F] |
| fr_sulfide | Number of thioether | [S&X2](-[#6])-C |
| fr_SH | Number of thiol groups | [S&H1] |
| fr_C_S | Number of thiocarbonyl | C=[S&X1] |
| fr_sulfone | Number of sulfone groups | S(=,-[O&X1;+0,-])(=,-[O&X1;+0,-])(-[#6])-[#6] |
| fr_sulfonamd | Number of sulfonamides | N-S(=,-[O&X1;+0,-])(=,-[O&X1;+0,-])-[#6] |
| fr_prisulfonamd | Number of primary sulfonamides | [N&H2]-S(=,-[O&X1&+0&-])(=,-[O&X1&+0&-])-[#6] |
| fr_barbitur | Number of barbiturate groups | C1C(=O)NC(=O)NC1=O |
| fr_urea | Number of urea groups | C(=O)(-N)-N |
| fr_term_acetylene | Number of terminal acetylenes | C#[C&H1] |
| fr_imidazole | Number of imidazole rings | n1cncc1 |
| fr_furan | Number of furan rings | o1cccc1 |
| fr_thiophene | Number of thiophene rings | s1cccc1 |
| fr_thiazole | Number of thiazole rings | c1scnc1 |
| fr_oxazole | Number of oxazole rings | c1ocnc1 |
| fr_pyridine | Number of pyridine rings | n1ccccc1 |
| fr_piperdine | Number of piperdine rings | N1CCCCC1 |
| fr_piperzine | Number of piperzine rings | N1CCNCC1 |
| fr_morpholine | Number of morpholine rings | O1CCNCC1 |
| fr_lactam | Number of beta lactams | N1C(=O)CC1 |
| fr_lactone | Number of cyclic esters (lactones) | [C&R1](=O)[O&R1][C&R1] |
| fr_tetrazole | Number of tetrazole rings | c1nnnn1 |
| fr_epoxide | Number of epoxide rings | O1CC1 |
| fr_unbrch_alkane | Number of unbranched alkanes of at least 4 members (excludes halogenated alkanes) | [R0&D2][R0&D2][R0&D2][R0&D2] |
| fr_bicyclic | Bicyclic | [R2][R2] |
| fr_benzene | Number of benzene rings | c1ccccc1 |
| fr_phos_acid | Number of phosphoric acid groups | [$(P(=[O&X1])([$([O&X2&H1]),$([O&X1&-]),$([O&X2]P)])([$([O&X2&H1]),$([O&X1&-]),$([O&X2]P)])[$([O&X2&H1]),$([O&X1&-]),$([O&X2]P)]),$([P&+]([O&X1&-])([$([O&X2&H1]),$([O&X1&-]),$([O&X2]P)])([$([O&X2&H1]),$([O&X1&-]),$([O&X2]P)])[$([O&X2&H1]),$([O&X1&-]),$([O&X2]P)])] |
| fr_phos_ester | Number of phosphoric ester groups | [$(P(=[O&X1])([O&X2][#6])([$([O&X2&H1]),$([O&X1&-]),$([O&X2][#6])])[$([O&X2&H1]),$([O&X1&-]),$([O&X2][#6]),$([O&X2]P)]),$([P&+]([O&X1&-])([O&X2][#6])([$([O&X2&H1]),$([O&X1&-]),$([O&X2][#6])])[$([O&X2&H1]),$([O&X1&-]),$([O&X2][#6]),$([O&X2]P)])] |
| fr_nitro_arom | Number of nitro benzene ring substituents | [$(c1(-[$([N&X3](=O)=O),$([N&X3&+](=O)[O&-])])ccccc1)] |
| fr_nitro_arom_nonortho | Number of non-ortho nitro benzene ring substituents | [$(c1(-[$([N&X3](=O)=O),$([N&X3&+](=O)[O&-])])ccccc1)&!$(cc-&!:*)] |
| fr_dihydropyridine | Number of dihydropyridines | [$([N&X3&H1]1-C=C-C-C=C1),$([N&v3]1=C-C-C=C-C1),$([N&v3]1=C-C=C-C-C1),$([N&X3&H1]1-C-C=C-C=C1)] |
| fr_phenol_noOrthoHbond | Number of phenolic OH excluding ortho intramolecular Hbond substituents | [$(c1(-[O&X2&H1])ccccc1)&!$(cc-&!:[C&H2]-[O&X2&H1])&!$(cc-&!:C(=O)[O;H1,-])&!$(cc-&!:C(=O)-[N&H2])] |
| fr_Al_OH_noTert | Number of aliphatic hydroxyl groups excluding tert-OH | [$(C-[O&X2&H1])&!$([C&X3](-[O&X2&H1])=[O&X1])&!$([C&D4]-[O&X2&H1])] |
| fr_benzodiazepine | Number of benzodiazepines with no additional fused rings | [c&R2]12[c&R1][c&R1][c&R1][c&R1][c&R2]1[N&R1][C&R1][C&R1][N&R1]=[C&R1]2 |
| fr_para_hydroxylation | Number of para-hydroxylation sites | [$([c&H1]1[c&H1]cc(c[c&H1]1)~[$([#8,$([#8]~[H1,c,C])])]),$([c&H1]1[c&H1]cc(c[c&H1]1)~[$([#7&X3,$([#7](~[H1,c,C])~[H1,c,C])])]),$([c&H1]1[c&H1]cc(c[c&H1]1)-&!:[$([N&X3&H1,$(NC(=O)[H1,c,C])])])] |
| fr_allylic_oxid | Number of allylic oxidation sites excluding steroid dienone | [$(C=C-C)&!$(C=C-C-[N,O,S])&!$(C=C-C-C-[N,O])&!$(C12=CC(=O)CCC1C1C(C3C(CCC3)CC1)CC2)] |
| fr_aryl_methyl | Number of aryl methyl sites for hydroxylation | [$(a-[C&H3]),$(a-[C&H2]-[C&H3]),$(a-[C&H2]-[C&H2]~[!N&!O]);!$(a(:a!:*):a!:*)] |
| fr_Ndealkylation1 | Number of XCCNR groups | [$(N(-[C&H3])-C-[$(C~O),$(C-a),$(C-N),$(C=C)]),$(N(-[C&H2][C&H3])-C-[$(C~O),$(C-a),$(C-N),$(C=C)])] |
| fr_Ndealkylation2 | Number of tert-alicyclic amines (no heteroatoms, not quinine-like bridged N) | [$([N&R1]1(-C)CCC1),$([N&R1]1(-C)CCCC1),$([N&R1]1(-C)CCCCC1),$([N&R1]1(-C)CCCCCC1),$([N&R1]1(-C)CCCCCCC1)] |
| fr_alkyl_carbamate | Number of alkyl carbamates (subject to hydrolysis) | C[N&H1]C(=O)OC |
| fr_ketone_Toplis | Number of ketones excluding diaryl, a,b-unsat. dienones, heteroatom on Calpha | [$([C&X3](=[O&X1])(C)[c,C])&!$([C&X3](=[O&X1])([C&H1]=C)[c,C])] |
| fr_ArN | Number of N functional groups attached to aromatics | [$(a-[N&X3&H2]),$(a-[N&H1][N&H2]),$(a-C(=[O&X1])[N&H1][N&H2]),$(a-C(=[N&H1])[N&H2])] |
| fr_HOCCN | Number of C(OH)CCN-Ctert-alkyl or C(OH)CCNcyclic | [$([O&X2&H1][C&X4][C&X4&H2][N&X3&R1]),$([O&H1][C&X4][C&X4&H2][N&X3][C&X4](C)(C)C)] |